\documentclass[letterpaper, 10 pt, conference]{ieeeconf}  
\IEEEoverridecommandlockouts                              
\usepackage{graphics}    
\usepackage{times}       
\usepackage{amsmath}     
\usepackage{amssymb}     
\usepackage{graphicx}
\usepackage{algorithm}
\usepackage[noend]{algpseudocode}
\usepackage{booktabs}
\usepackage{subfigure}
\usepackage{rotating}

\usepackage[pagebackref=true,breaklinks=true,letterpaper=true,colorlinks,bookmarks=false]{hyperref}

\usepackage[font=small]{caption}

\def\reffig#1{Fig.~\ref{#1}}
\def\reftab#1{Tab.~\ref{#1}}

\makeatletter

\usepackage{xspace}
\DeclareRobustCommand\onedot{\futurelet\@let@token\@onedot}
\def\@onedot{\ifx\@let@token.\else.\null\fi\xspace}

\def\ie{i.e\onedot} 
\def\cf{cf\onedot}

\def\etal{\emph{et al}\onedot}
\makeatother

\usepackage{array}
\newcolumntype{L}[1]{>{\raggedright\let\newline\\\arraybackslash\hspace{0pt}}m{#1}}
\newcolumntype{C}[1]{>{\centering\let\newline\\\arraybackslash\hspace{0pt}}m{#1}}
\newcolumntype{R}[1]{>{\raggedleft\let\newline\\\arraybackslash\hspace{0pt}}m{#1}}

\newcommand\set[1]{\mathcal{#1}}
\renewcommand\vec[1]{\mathbf{#1}}

\title{\LARGE \bf A Benchmark for LiDAR-based Panoptic Segmentation based on KITTI}

\author{Jens Behley \and Andres Milioto \and Cyrill Stachniss
  \thanks{All authors are with the University of Bonn, Germany. }%
  \thanks{This work has partially been funded by the Deutsche Forschungsgemeinschaft (DFG, German Research Foundation) under grant number BE 5996/1-1 and under Germany's Excellence Strategy, EXC-2070 - 390732324 - PhenoRob.
  }%
}

\begin{document}
\maketitle
\thispagestyle{empty}
\pagestyle{empty}

\begin{abstract}
  Panoptic segmentation is the recently introduced task~\cite{kirillov2019cvpr-ps} that tackles semantic segmentation and instance segmentation jointly. In this paper, we present an extension of SemanticKITTI~\cite{behley2019iccv}, which is a large-scale dataset providing dense point-wise semantic labels for all sequences of the KITTI Odometry Benchmark~\cite{geiger2012cvpr}, for training and evaluation of laser-based panoptic segmentation. We provide the data and discuss the processing steps needed to enrich a given semantic annotation with temporally consistent instance information, \ie, instance information that supplements the semantic labels and identifies the same instance over sequences of LiDAR point clouds. Additionally, we present two strong baselines that combine state-of-the-art LiDAR-based semantic segmentation approaches with a state-of-the-art detector enriching the segmentation with instance information and that allow other researchers to compare their approaches against. We hope that our extension of SemanticKITTI with strong baselines enables the creation of novel algorithms for LiDAR-based panoptic segmentation as much as it has for the original semantic segmentation and semantic scene completion tasks. Data, code, and an online evaluation using a hidden test set will be published on \url{http://semantic-kitti.org}.
  \vspace{0.5cm}
\end{abstract}

\section{Introduction}
\label{sec:intro}

Fine-grained scene understanding is a pre-requisite for truly autonomous systems.
This encompasses the type of surfaces, but also identifying individual objects.
The former is often designated as \emph{stuff} and the latter as \emph{things}~\cite{kirillov2019cvpr-ps}.
Only both sources of information together enable autonomous systems to reason about the drivability of surfaces, the type of objects and obstacles, and possibly the intent of other agents in the vicinity.

\begin{figure}[t]
  \centering
  \includegraphics[width=0.98\linewidth]{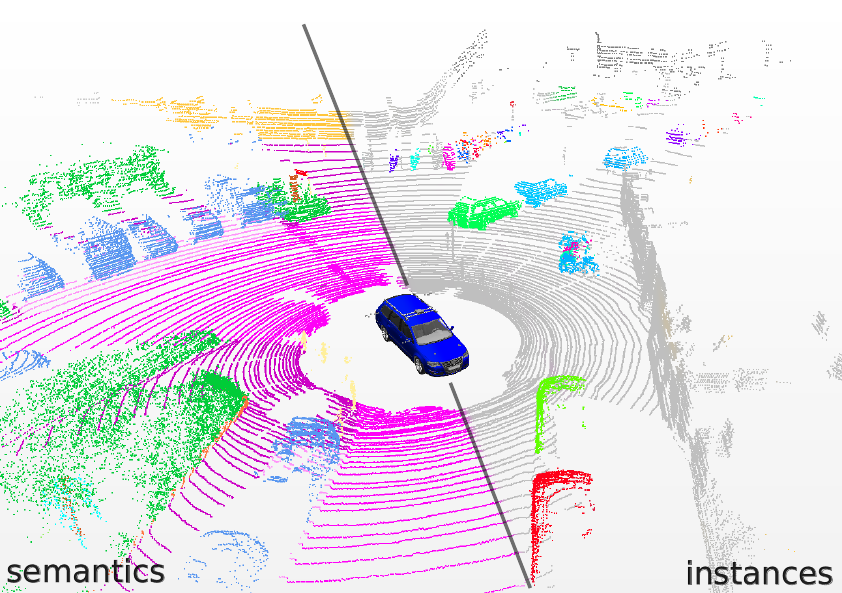}
  \caption{Using the semantic segmentation (left part) and the point-accurate instance annotations for traffic participants (right part), we provide a benchmark for panoptic segmentation~\cite{kirillov2019cvpr-ps} using three-dimensional LiDAR point clouds.
    Our work extends the SemanticKITTI~\cite{behley2019iccv} dataset, which is based on the KITTI Vision Benchmark~\cite{geiger2012cvpr}.}
  \label{fig:motivation}
\end{figure}

Assigning to each individual pixel or point a semantic label is called semantic segmentation, while the identification and separation of individual objects is called instance segmentation.
These tasks are usually solved in isolation, but an increasing number of methods were recently developed that solve both jointly using either images~\cite{kirillov2019cvpr-ps,porzi2019cvpr,xiong2019cvpr-uaup} or RGB-D data~\cite{pham2019cvpr,hou2019cvpr-siso}.
This development was mainly driven by the availability of a metric~\cite{kirillov2019cvpr-ps} and the swift adaption of the task in different popular semantic segmentation datasets, such as Cityscapes~\cite{cordts2016cvpr}, Microsoft's COCO~\cite{lin2014eccv}, and Mapillary Vistas~\cite{neuhold2017iccv}.
While semantic segmentation will still be relevant in the future, we expect that instance segmentation will be soon replaced and subsumed by the panoptic segmentation task, as a part of a panoptic segmentation framework.

In this paper, we present an extension of the SemanticKITTI dataset~\cite{behley2019iccv} providing the necessary annotations to evaluate panoptic segmentation on automotive LiDAR scans.
\reffig{fig:motivation} shows an example of the provided instance annotation for all traffic participants, \ie, vehicles, pedestrians, and cyclists.
To ease the generation of instance information with provided semantic segmentation of the LiDAR point clouds, we first generate for static and non-static objects instance information using grid-based clustering~\cite{behley2013iros} and distance-based clustering approach.
However, this clustering leads often to over- or under-segmentation, which must be manually corrected using our point labeling tool.
Furthermore, we provide two baseline approaches that combine state-of-the-art semantic segmentation with state-of-the-art object detection methods.

In summary, our contributions are as follows:
\begin{itemize}
  \item We provide temporally-consistent instance annotations for all traffic participants including vehicles, pedestrians, bicyclists, and motorcyclists for the KITTI Odometry Benchmark.
  \item We present two strong baseline approaches combining current state-of-the-art semantic segmentation and a state-of-the-art 3D object detector.
  \item We enable an online evaluation platform for approaches to LiDAR-based panoptic segmentation using a hidden test set.
\end{itemize}


\begin{figure*}[t]
  \vspace{2.5mm} 
  \centering
  \includegraphics[trim=40 0 0 100,clip,width=0.49\linewidth]{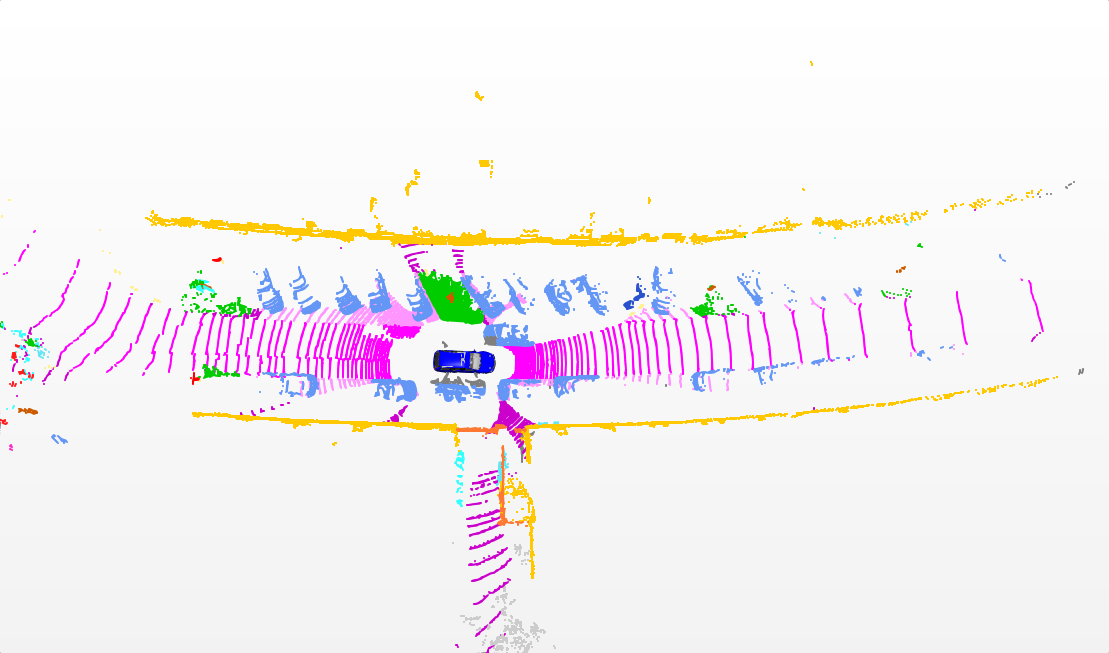}\hspace{0.08cm}
  \includegraphics[trim=40 0 0 100,clip,width=0.49\linewidth]{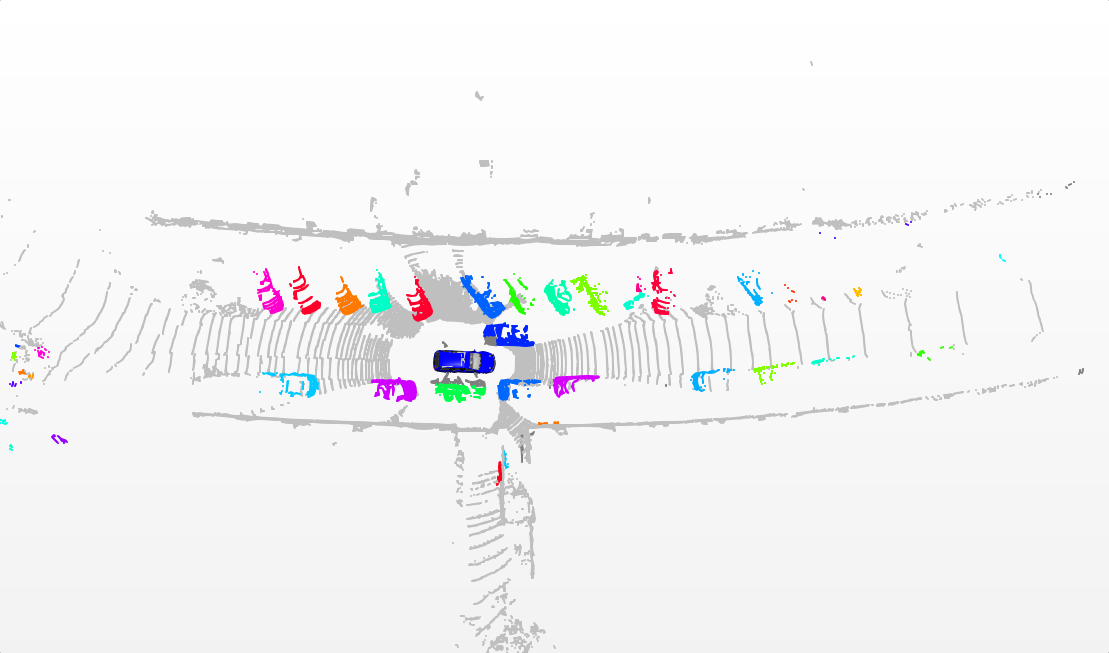}\\[2mm]
  \includegraphics[trim=40 100 0 100,clip,width=0.49\linewidth]{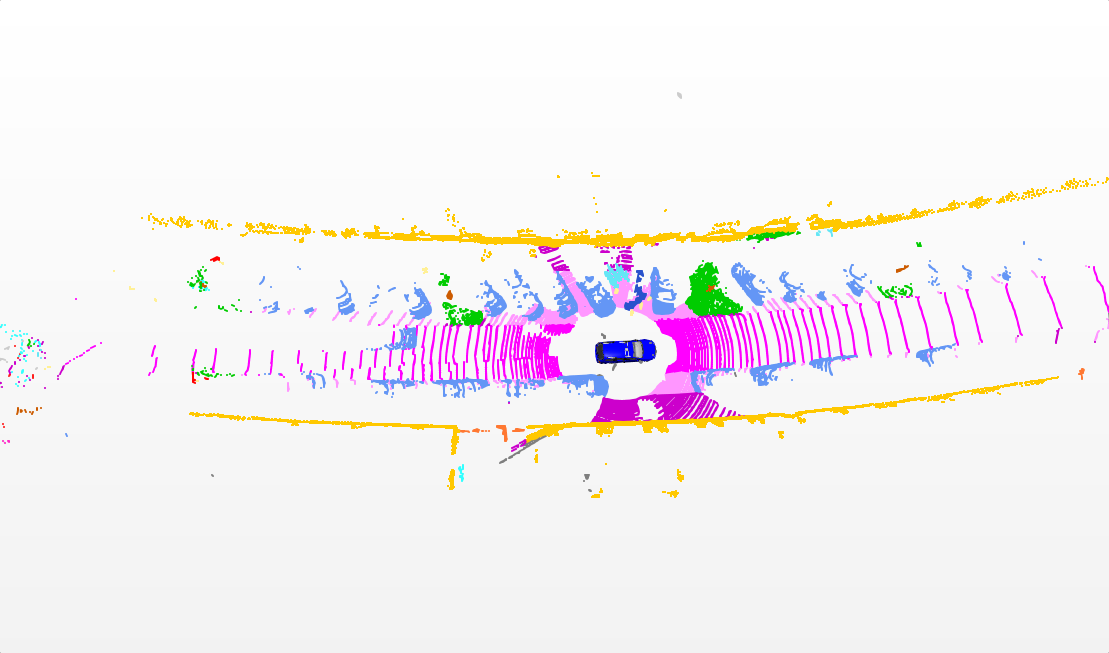}\hspace{0.08cm}
  \includegraphics[trim=40 100 0 100,clip,width=0.49\linewidth]{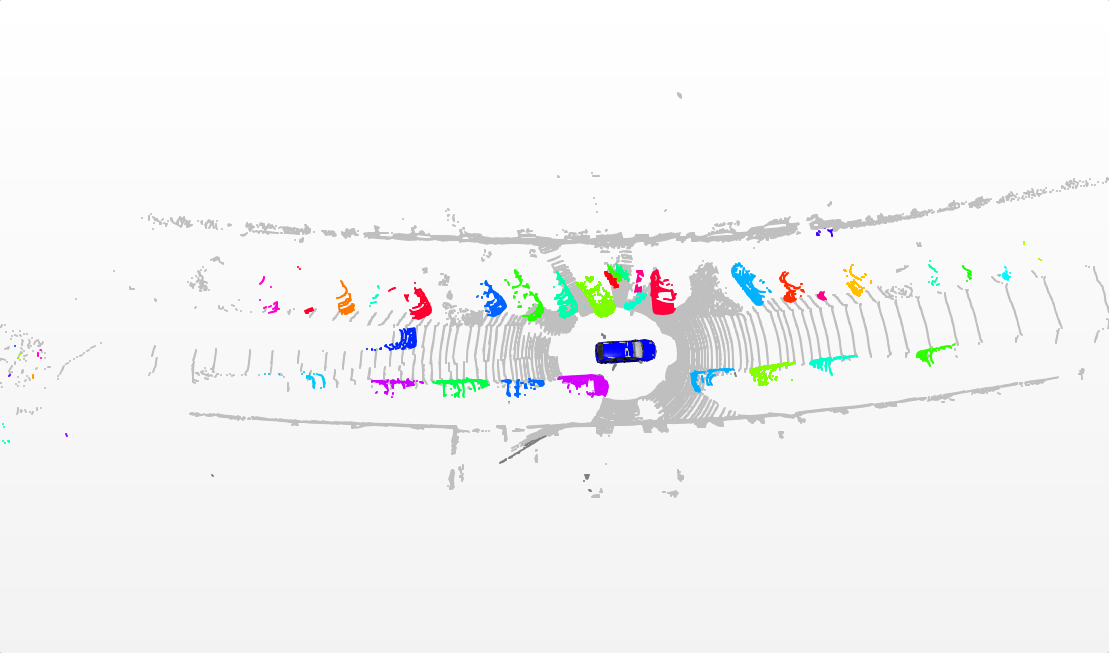}
  \caption{Qualitative example of the instance annotation over a sequence of scans: on the left is the semantic annotation and on the right is the instance annotation shown.
    Note, same colors at different timestamps correspond to the same instance id. Best viewed in color.}
  \label{fig:annotation_examples}
\end{figure*}

\section{Related Work}
\label{sec:related}

\textbf{Panoptic segmentation.} Shortly after Kirillov \etal~\cite{kirillov2019cvpr-ps} proposed panoptic segmentation and a metric to measure the performance of approaches providing such labels, the established datasets for semantic segmentation of \emph{image data}, \ie, Cityscapes~\cite{cordts2016cvpr}, Microsoft's Common Objects in Context (COCO)~\cite{lin2014eccv}, and Mapillary's Vistas~\cite{neuhold2017iccv} adopted the metric and added an evaluation for this task.
The last version of the Joint COCO and Mapillary Recognition Challenge workshop at ICCV also featured a panoptic segmentation track.

Due to the availability of the data, we witnessed a wide adoption and interest for panoptic segmentation in the computer vision community~\cite{cheng2019iccvw,kirillov2019cvpr-ps,liu2019cvpr-aenf,pham2019cvpr,porzi2019cvpr,xiong2019cvpr-uaup}.
While there have also been approaches for RGB-D data~\cite{hou2019cvpr-siso,pham2019cvpr}, there are currently no approaches available that operate on LiDAR data, probably because there was no annotated data available that provides point-wise semantic labels \emph{and} instance information at the same time.

\textbf{Datasets.} Recently, almost all major self-driving car companies release datasets providing besides camera also LiDAR data, including Waymo~\cite{sun2019arxiv}, Lyft~\cite{kesten2019misc}, Audi~\cite{geyer2019misc}, Argo~\cite{chang2019cvpr}, and Aptiv~\cite{caesar2019arxiv}.
While all datasets provide also the annotations for object instances by bounding boxes, only a few datasets provide point-wise semantic annotation~\cite{behley2019iccv, geyer2019misc} needed to evaluate panoptic segmentation for LiDAR.

SemanticKITTI~\cite{behley2019iccv} is a dataset based on the KITTI Vision Benchmark~\cite{geiger2012cvpr}, which might not show the diversity of different inner cities traffic and weather conditions, but still provides unparalleled long sequences showing a wide variety of different environments and driving situations.
Our annotations with point-wise labels for the full $360^\circ$ field-of-view provide labels for 28 classes including labels distinguishing moving and non-moving objects.
By providing now instance annotations together with an online evaluation on a hidden test set, we close the gap to the aforementioned established image-based dataset and provide means to evaluate panoptic segmentation using an automotive LiDAR.
We hope that the availability of labeled LiDAR scans for panoptic segmentation opens the door for more research in the direction of LiDAR-based panoptic segmentation.

\section{Dataset}
\label{sec:main}

In this section, we introduce the provided dataset and discuss the annotation process to extract instance information from a given semantic segmentation in a semi-automatic fashion with acceptable manual labeling effort to adjust for wrong over- and under-segmentations.

\reffig{fig:annotation_examples} shows a qualitative example of the annotation provided by our dataset.
The left part of the figure shows the semantic segmentation of our SematicKITTI dataset, which we use to determine the instance annotation.
The right side depicts the temporally consistent annotations,  where different colors correspond to individual instances.
Note, that same colored instances in the top row and the bottom row of this figure correspond also to the same instance getting the same instance ID.

\begin{figure}
  {\includegraphics[trim=0 200 0 0,clip,width=\linewidth]{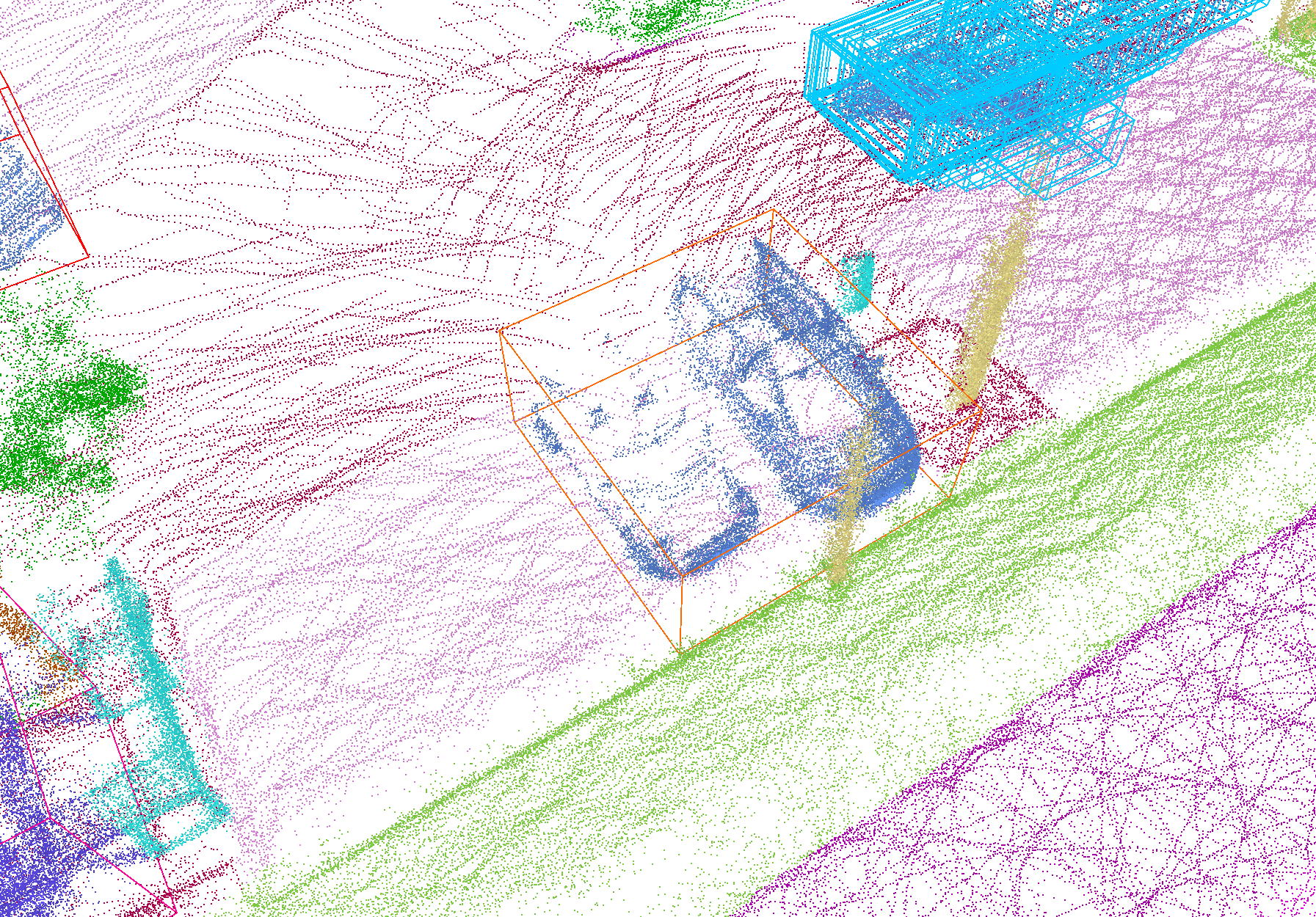}}\\[2mm]
  {\includegraphics[trim=0 200 0 0,clip,width=\linewidth]{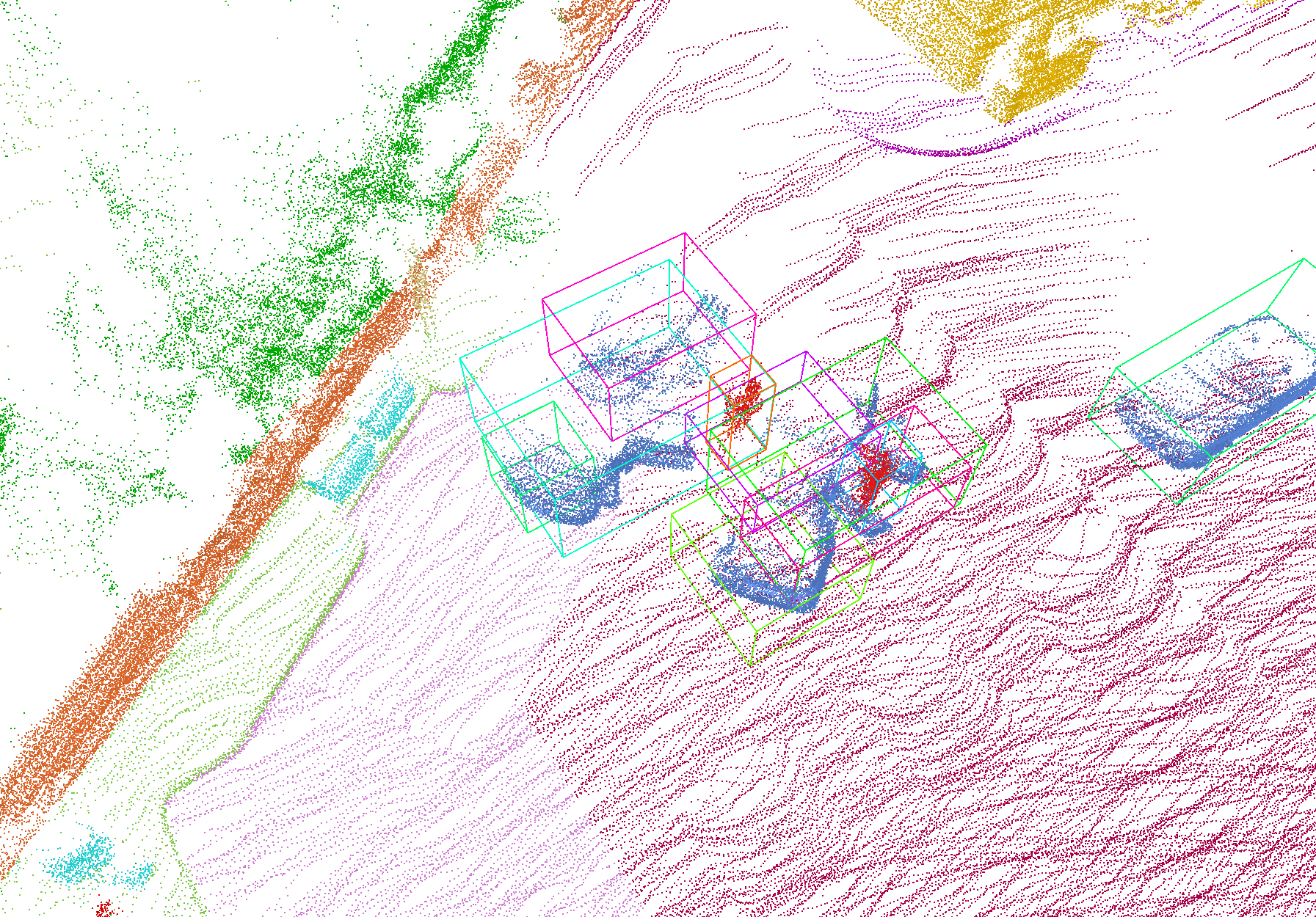}}
  \caption{Example of under- (top) and over-segmentation (bottom) generated by our semi-automated clustering approach.}
  \label{fig:example_over_under_segmentation}
\end{figure}

\subsection{Annotation Process}

For annotation of the instances, we employ a semi-automatic process using different strategies to generate temporally consistent instance annotation.
Our goal is to label the same instance through the whole sequence with the same instance ID -- even for instances that move.
For static objects, the data association can be simply performed by considering the location of the segment after performing a pose correction using a SLAM system~\cite{behley2018rss}.
For moving objects, we have to account for the motion of the object as well as the motion of the sensor at the same time.
 
Overall, the SemanticKITTI dataset~\cite{behley2019iccv} provides 28 classes (including 6 classes to distinguish moving from non-moving classes) from which we select the traffic participants as \emph{thing} classes for the panoptic segmentation, \ie, car, truck, other-vehicle, motorcycle, bicycle, person, bicyclist, and motorcyclist.
The remaining classes are \emph{stuff} classes for the panoptic segmentation, \ie, road, sidewalk, parking, other-ground, building, vegetation, trunk, terrain, fence, pole, and traffic-sign.

For static thing classes, we first cluster all points for each individual class using a fast grid-based segmentation approach~\cite{behley2013iros} to handle the large number of points efficiently.
We then split the aggregated point cloud into tiles of size $100\,$m by $100\,$m using the pose information by our SLAM system~\cite{behley2018rss}.
For each tile, we use a two-dimensional grid with cell size $0.1\,$m by $0.1\,$m, which allows us often to separate even close parking cars.
Next, all points are inserted into the corresponding grid cells using their~$x$ and~$y$-coordinates.
Finally, only grid cells with points exceeding a height threshold $\Delta > 0.5$\,m are combined using a flood fill algorithm to combine neighboring grid cells into segments.

For moving thing classes, we generate clusters for each scan individually using a distance-based clustering as this provided more reliable results and could be also used to associate instances between consecutive scans.
First, we search for each point in a radius of $0.5$\,m for the nearest neighbor and cluster points together that share neighbors.
To find associations with the previous $4$ scans, we use a slightly larger radius of $1.0$\,m to find neighbors between two different timestamps.
If we find enough neighbors with the previous segments at different timestamps, we associate them together and assign the same instance ID.

The described clustering leads inevitably to over- and under-segmentation (\cf \reffig{fig:example_over_under_segmentation}), but also to wrong or missing associations between consecutive timestamps.
We correct these issues manually using an own point labeling tool, which provides tools to create, join, and split instances.
Overall, the manual correction for all $22$ sequences took roughly $70$\,h of additional labor.

\begin{figure}

  \centering
  {\includegraphics[]{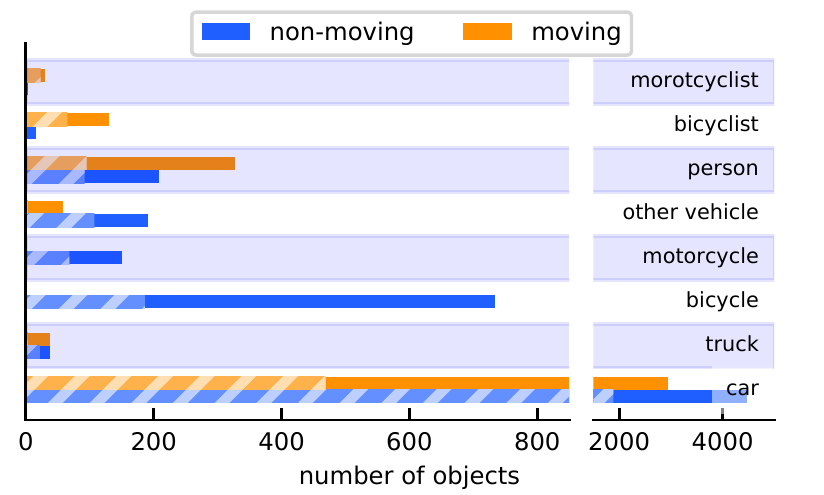}}\\[2mm]
  {\includegraphics[]{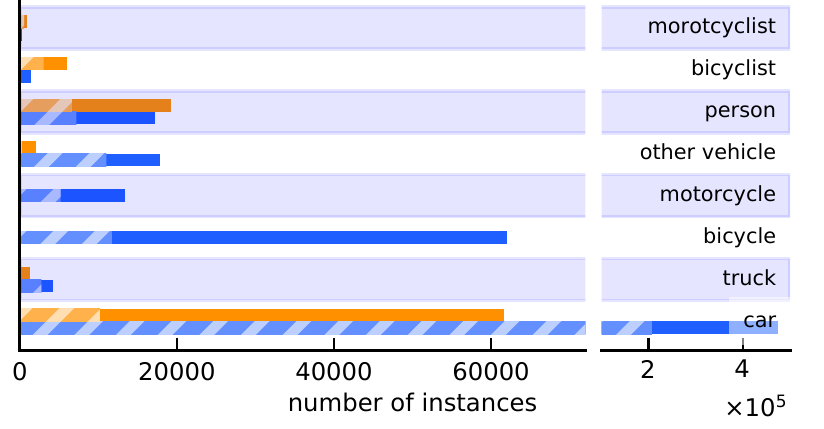}}
  \caption{Top: number of (sequence-wise) objects. Bottom: number of (scan-wise) instances. The hashed bars correspond to the training data. The large number of scan-wise annotations in relation to the number of objects indicates that many objects are seen over an extended period of time.}
  \label{fig:instance_statistics}
\end{figure}

\subsection{Statistics}

\reffig{fig:instance_statistics} provides an overview of the number of instances and the actual number of bounding boxes per class.
We show in the upper part the sequence-wise counts of instance annotations, \ie, we count each object only once even if it is seen multiple times by the sensor.
The lower part of the figure shows the accumulated scan-wise counts of instances, where we count the instances without considering the temporally consistent instance ID.

The bulk of the instances correspond to cars, which are naturally occurring in city-like environments and also correspond to the normal statistics in autonomous driving scenarios.
Usually, an autonomous car will also encounter some classes far fewer than other classes or situations. They are usually denoted as the `long tail' problem, referring to the underrepresented entities in a given distribution.
This adds complexity to the task, since panoptic segmentation approaches, which are designed to tackle this scenario must be able to deal with such skewed class distributions.

\section{Two-Stage Baseline Approaches}
\label{sec:twostage}

We provide two baseline approaches with this dataset. Our baseline approaches are a combination of the current state-of-the-art semantic segmentation approaches on SemanticKITTI, namely KPConv~\cite{thomas2019iccv} and RangeNet++~\cite{milioto2019iros}, paired with a state-of-the-art object detector, namely PointPillars~\cite{lang2019cvpr}, providing instance-level information.

To this end, we use the oriented bounding boxes of the object detector, \ie, bounding boxes for \emph{cars}, \emph{pedestrians}, and \emph{cyclists} trained on the KITTI detections benchmark~\cite{geiger2012cvpr}, to determine the instance ID for points inside the bounding boxes.
By combining the predictions of the semantic segmentation and assigning the instance ID of each bounding box to each point inside of it, we obtain a panoptic segmentation.
Note that we only assign instance IDs to points from the \emph{thing} classes, \ie, points under a \emph{car} classified as \emph{road} or \emph{parking} are not assigned an instance ID.

For the baseline, we used pre-trained models or publicly available predictions for KPConv~\cite{thomas2019iccv} and RangeNet++~\cite{milioto2019iros}, which were trained on SemanticKITTI.
PointPillars had to be trained from scratch using the provided implementation\footnote{See the GitHub repository at \mbox{\url{https://git.io/Je25l}}.}, modifying the configuration of the object detector such that it provides region proposals and bounding boxes for the full 360 degree field-of-view of the LiDAR sensor.

These networks were run independently for semantic segmentation and object detection and then merged to generate a panoptic segmentation.
Neither approach can, therefore, run at the frame rate of the LiDAR, \ie, $10$ Hz, and thus having computational budgets that are not suitable in an autonomous car.
Furthermore, the PointPillars detector~\cite{lang2019cvpr} requires training separate networks, one for the class \emph{car} and one for \emph{pedestrian} combined with \emph{cyclist}, which accentuates the problem further.
We provide an evaluation of the performance of these approaches including runtime information in the experimental section of this paper.

\begin{figure}
  \centering
  \includegraphics[trim=0 30 0 30,clip,width=\linewidth]{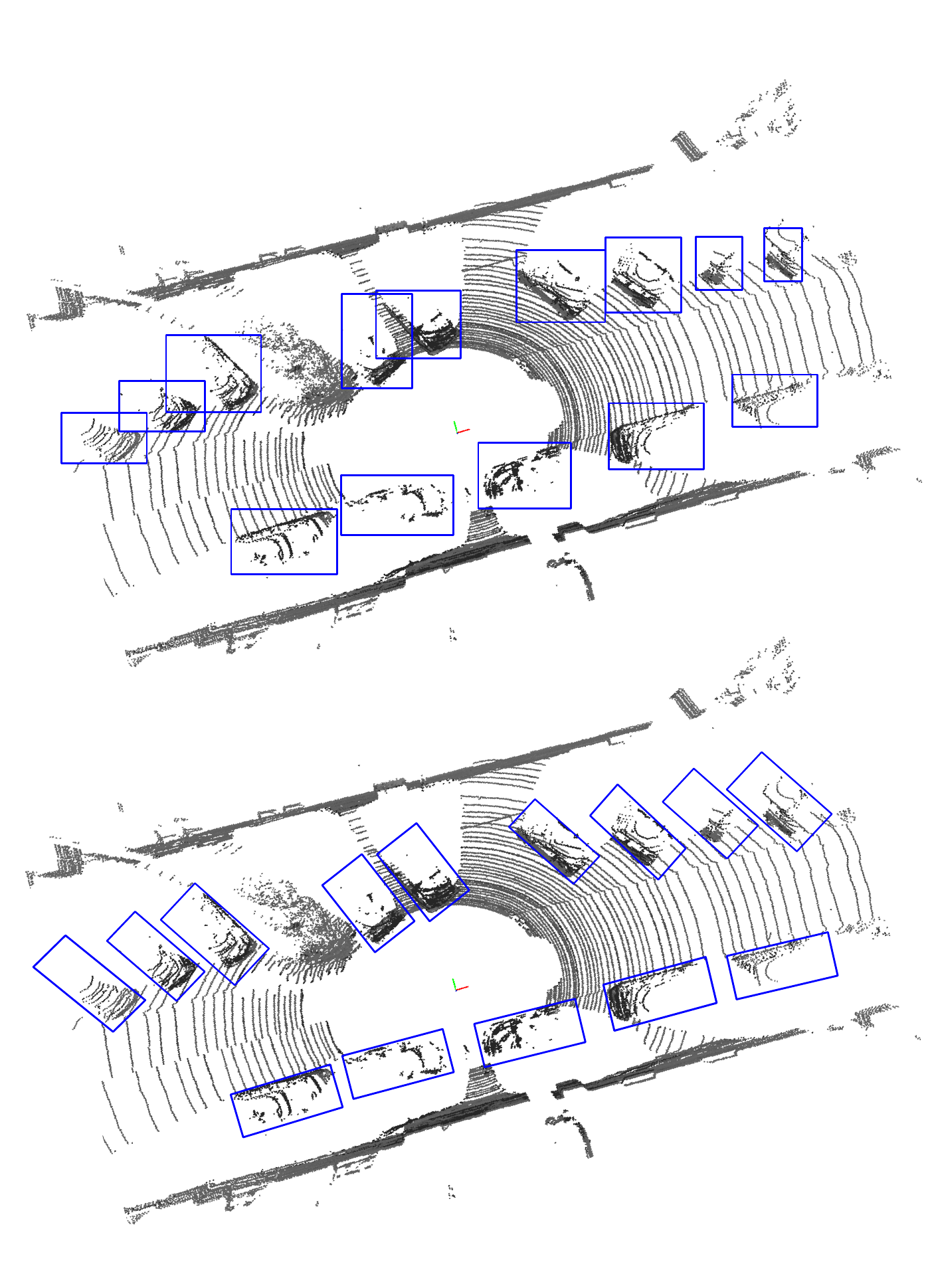}
  \caption{Overlapping of axis-aligned bounding boxes and therefore wrong or ambiguous assignment of points inside bounding boxes (top).
    With oriented bounding boxes this ambiguity due to overlapping bounding boxes does not occur (bottom).}
  \label{fig:obb_vs_aabb}
\end{figure}

Note that the decision for using an object detector providing oriented bounding boxes was made to minimize the negative effect of axis-aligned bounding boxes, which would lead to large overlaps between cars parking near to each other, see also \reffig{fig:obb_vs_aabb} for an example. Thus, oriented bounding boxes lead to more accurate instance annotations in the depicted case.

\section{Experiments}
\label{sec:exp}

Before we discuss details of the baseline implementations and the results of our baseline approaches, we shortly provide a summary of the panoptic segmentation metric.

\subsection{Evaluation Metric}

In panoptic segmentation, each point $\mathbf{p}_i$ not only caries a class label $y_i \in \mathcal{Y}$, where $|\mathcal{Y}|$ is the number of classes, but also can have an instance ID~$n_i$, where~$n_i = 0$ denotes no specific instance.

To measure the quality of this joint assignment, we briefly recapitulate the recently proposed panoptic quality~(PQ) metric~\cite{kirillov2019cvpr-ps}.
Let $\set{S}, \hat{\set{S}}$ denote segments, \ie, sets of points in our specific case, sharing an class and instance ID. Here, we assume that the \emph{stuff} classes, e.g., vegetation, simply get instance ID~$n_i = 0$ corresponding to no specific instance assigned.

Furthermore, let IoU($\set{S}, \hat{\set{S}}$) = $(\set{S} \cap \hat{\set{S}})  \cdot (\set{S} \cup \hat{\set{S}})^{-1} $  denote the intersection-over-union of these two sets.
Let the set of true positive matches~$\text{TP}_c$ be the pairs of predicted segments~$\hat{\set{S}}$ that overlap at least with $0.5$ IoU with a ground truth segment~$\set{S}$, $\text{TP}_c = \{(\set{S}, \hat{\set{S}}) \mid  \text{IoU}(\set{S}, \hat{\set{S}}) > 0.5\}$.
Likewise, let~$\text{FP}_c$ the set of unmatched predicted segments~$\hat{\set{S}}$ and~$\text{FN}_c$  the set of unmatched ground truth segments ${\set{S}}$.

With the above definitions, the class-wise PQ$_c$ is  given by
\begin{align}
  \text{PQ}_c & = \frac{ \sum_{(\set{S}, \hat{\set{S}}) \in \text{TP}_c} \text{IoU}(\set{S}, \hat{\set{S}})}{|\text{TP}_c| + \frac{1}{2}|\text{FP}_c|  + \frac{1}{2}|\text{FN}_c|}.
\end{align}

The panoptic quality metric is computed for each class independently and averaged over all classes, which makes the metric insensitive to class imbalance~\cite{kirillov2019cvpr-ps}, \ie,
\begin{align}
  \text{PQ} & = \frac{1}{|\mathcal{Y}|} \sum_{c \in \set{Y}} \text{PQ}_c. \label{eq:panoptic_quality}
\end{align}

For images, Porzi~\etal~\cite{porzi2019cvpr} proposed to alter the metric to account for \emph{stuff} classes having only a single segment since no pixels (or, in our case, points) have an instance ID.
In such a case, the IoU-based criterion could often lead to an unmatched prediction.
To account for \emph{stuff} classes, Porzi \etal use
\begin{align}
  \text{PQ}^{\dagger}_c & = \left\{\begin{array}{cl}  \text{IoU}(\set{S}, \hat{\set{S}}) & , \text{if c is a \emph{stuff} class} \\ \text{PQ}_c &, \text{otherwise.} \end{array}\right.
\end{align}
Consequently, we denote by PQ$^{\dagger}$ the average over the class-wise modified $\text{PQ}^{\dagger}_c$ as defined in \eqref{eq:panoptic_quality}.

Furthermore, the quality of the semantic segmentation is also measured using the mean intersection-over-union (mIoU), which also enables the comparison with other approaches in the semantic segmentation benchmark.
This metric is defined as follows:
\begin{align}
  \text{mIoU} & = \frac{1}{|\mathcal{Y}|}~\sum_{c \in \mathcal{Y}}\frac{|\{i \mid y_i = c\} \cap \{j \mid \hat{y}_j = c\}|}{|\{i \mid y_i = c\} \cup \{j \mid \hat{y}_j = c\}|},
\end{align}
where $y_i$ corresponds to the ground truth label of point $\vec{p}_i$ and $\hat{y}_i$ to the prediction.

\subsection{Basline Parameters, Training and Inference Details}

In this section, we provide more details on the training and inference of the two-stage baselines.
We, furthermore, provide details on the modifications needed to use the models on the SemanticKITTI~\cite{behley2019iccv} benchmark, which requires to use full point clouds of a single turn for training and inference.
We will provide code for merging the predictions and configuration files to enable the reproduction of our results.

\textbf{KPConv by Thomas \etal~\cite{thomas2019iccv}.}
For scene classification, Thomas \etal~\cite{thomas2019iccv}, code-url\footnote{\url{https://github.com/HuguesTHOMAS/KPConv}}, extract~$10$ overlapping spheres of~ $10\,$m radius, subsample the point clouds with a voxel grid of size~$10\,$cm, and drop randomly points if there were more then~$15,000$ points inside the sphere.
To aggregate predictions, they perform majority vote on the overlapping parts of the predictions.
Overall, this achieves state-of-the-art single scan performance of $58.5$ mIoU and performs better than taking a subsampled single point cloud with mIoU~$56.6$ (subsampling with voxel grid of resolution $0.1$\,m).

\textbf{RangeNet++ by Milioto \etal~\cite{milioto2019iros}}.
Here, we directly use the predictions available in our repository\footnote{\url{https://github.com/PRBonn/lidar-bonnetal}}, which are also provided for the test set.
RangeNet++ uses a range image of size~$2048 \times 64$ for training and inferences, which is then upsampled to the complete point cloud by using nearest neighbors.
To remove artifacts from the reprojection, it applies a k-nearest neighbor filtering, which accounts for~$k$ neighbors in a certain range.

\textbf{PointPillars by Lang \etal~\cite{lang2019cvpr}}.
We used for training the implementation supported by the Point Pillar authors, code-url\footnote{\url{https://github.com/traveller59/second.pytorch}}. 
Since SemanticKITTI does not offer oriented bounding boxes, we use the 3D object detection part of KITTI Object Detection~\cite{geiger2012cvpr} for training.
The KITTI dataset was recorded with the same sensor and a similar environment, but there is no overlap between the point cloud sequences of the odometry and the detection benchmark.

\begin{table}[t]
  \centering
  \footnotesize{
    \begin{tabular}{L{2.0cm}C{1.5cm}C{1.5cm}C{1.5cm}}
      \toprule
      Class      & Easy    & Medium  & Hard    \\
      \midrule
      Car        & $86.30$ & $75.50$ & $69.44$ \\
      Pedestrian & $59.73$ & $54.98$ & $50.10$ \\
      Cyclist    & $76.82$ & $57.99$ & $54.35$ \\
      \bottomrule
    \end{tabular}
  }

  \caption{Validation set results (3d metric) of our retrained detector. Car is evaluated at $0.7$ IoU, and $0.5$ IoU otherwise.}
  \label{tab:validation_results}
  \vspace{-0.5cm}
\end{table}

\setlength\tabcolsep{5.5pt}\begin{table*}[t]
\centering\begin{tabular}{lc|cccc|ccc|ccc}
\toprule
Method & mIoU & PQ & PQ$^\dagger$ & RQ & SQ & PQ$^{\text{Th}}$ & RQ$^{\text{Th}}$ & SQ$^{\text{Th}}$ & PQ$^{\text{St}}$ & RQ$^{\text{St}}$ & SQ$^{\text{St}}$ \\ 
\midrule
KPConv \cite{thomas2019iccv} + PointPillars \cite{lang2019cvpr} & 58.8 & 44.5 & 52.5 & 54.4 & 80.0 & 32.7 & 38.7 & 81.5 & 53.1 & 65.9 & 79.0\\
RangeNet++ \cite{milioto2019iros} + PointPillars \cite{lang2019cvpr} & 52.4 & 37.1 & 45.9 & 47.0 & 75.9 & 20.2 & 25.2 & 75.2 & 49.3 & 62.8 & 76.5\\
\bottomrule
\end{tabular}
\caption{Comparison of test set results on SemanticKITTI using \emph{stuff}{\scriptsize (St)} and \emph{thing}{\scriptsize (Th)} \emph{classes}. All results in [\%].}
\label{tab:panoptic_classes}
\end{table*}

\setlength\tabcolsep{1.9pt}\begin{table*}[t]
\centering\footnotesize{
\begin{tabular}{lcccccccccccccccccccc}
\toprule
Method & mIoU & \begin{sideways}road\end{sideways} & \begin{sideways}sidewalk\end{sideways} & \begin{sideways}parking\end{sideways} & \begin{sideways}other ground\end{sideways} & \begin{sideways}building\end{sideways} & \begin{sideways}car\end{sideways} & \begin{sideways}truck\end{sideways} & \begin{sideways}bicycle\end{sideways} & \begin{sideways}motorcycle\end{sideways} & \begin{sideways}other-vehicle\end{sideways} & \begin{sideways}vegetation\end{sideways} & \begin{sideways}trunk\end{sideways} & \begin{sideways}terrain\end{sideways} & \begin{sideways}person\end{sideways} & \begin{sideways}bicyclist\end{sideways} & \begin{sideways}motorcyclist\end{sideways} & \begin{sideways}fence\end{sideways} & \begin{sideways}pole\end{sideways} & \begin{sideways}traffic sign\end{sideways} \\ 
\midrule
KPConv \cite{thomas2019iccv}/PointPillars \cite{lang2019cvpr} & 58.8  & 88.8 & 72.7 & 61.3 & 31.6 & 90.5 & 96.0 & 33.4 & 30.2 & 42.5 & 44.3 & 84.8 & 69.2 & 69.1 & 61.5 & 61.6 & 11.8 & 64.2 & 56.4 & 47.4\\
RangeNet++ \cite{milioto2019iros}/PointPillars \cite{lang2019cvpr} & 52.4  & 91.8 & 75.1 & 65.0 & 27.7 & 87.4 & 91.5 & 26.2 & 26.0 & 34.6 & 23.7 & 80.5 & 55.1 & 64.8 & 38.8 & 40.2 & 5.6 & 58.6 & 47.9 & 55.9\\
\bottomrule
\end{tabular}}
\caption{Detailed class-wise results of test set results on SemanticKITTI in intersection-over-union (IoU). All results in [\%].}
\label{tab:results_iou_class}
\end{table*}

\setlength\tabcolsep{1.9pt}\begin{table*}[t]
\centering\footnotesize{
\begin{tabular}{lcccccccccccccccccccc}
\toprule
Method & PQ & \begin{sideways}road\end{sideways} & \begin{sideways}sidewalk\end{sideways} & \begin{sideways}parking\end{sideways} & \begin{sideways}other ground\end{sideways} & \begin{sideways}building\end{sideways} & \begin{sideways}car\end{sideways} & \begin{sideways}truck\end{sideways} & \begin{sideways}bicycle\end{sideways} & \begin{sideways}motorcycle\end{sideways} & \begin{sideways}other-vehicle\end{sideways} & \begin{sideways}vegetation\end{sideways} & \begin{sideways}trunk\end{sideways} & \begin{sideways}terrain\end{sideways} & \begin{sideways}person\end{sideways} & \begin{sideways}bicyclist\end{sideways} & \begin{sideways}motorcyclist\end{sideways} & \begin{sideways}fence\end{sideways} & \begin{sideways}pole\end{sideways} & \begin{sideways}traffic sign\end{sideways} \\ 
\midrule
KPConv \cite{thomas2019iccv}/PointPillars \cite{lang2019cvpr} & 44.5  & 84.6 & 60.1 & 34.1 & 8.8 & 80.7 & 72.5 & 17.2 & 9.2 & 30.8 & 19.6 & 77.6 & 53.9 & 42.2 & 29.9 & 59.4 & 22.8 & 49.0 & 46.2 & 46.8\\
RangeNet++ \cite{milioto2019iros}/PointPillars \cite{lang2019cvpr} & 37.1  & 90.6 & 63.2 & 41.3 & 6.7 & 79.2 & 66.9 & 6.7 & 3.1 & 16.2 & 8.8 & 71.2 & 34.6 & 37.4 & 14.6 & 31.8 & 13.5 & 38.2 & 32.8 & 47.4\\
\bottomrule
\end{tabular}}
\caption{Detailed class-wise results of test set results on SemanticKITTI in panoptic quality (PQ) \cite{kirillov2019cvpr-ps}. All results in [\%].}
\label{tab:results_panoptic_class}
\end{table*}

\setlength\tabcolsep{1.9pt}\begin{table*}[t]
\centering\footnotesize{
\begin{tabular}{lcccccccccccccccccccc}
\toprule
Method & PQ$^\dagger$ & \begin{sideways}road\end{sideways} & \begin{sideways}sidewalk\end{sideways} & \begin{sideways}parking\end{sideways} & \begin{sideways}other ground\end{sideways} & \begin{sideways}building\end{sideways} & \begin{sideways}car\end{sideways} & \begin{sideways}truck\end{sideways} & \begin{sideways}bicycle\end{sideways} & \begin{sideways}motorcycle\end{sideways} & \begin{sideways}other-vehicle\end{sideways} & \begin{sideways}vegetation\end{sideways} & \begin{sideways}trunk\end{sideways} & \begin{sideways}terrain\end{sideways} & \begin{sideways}person\end{sideways} & \begin{sideways}bicyclist\end{sideways} & \begin{sideways}motorcyclist\end{sideways} & \begin{sideways}fence\end{sideways} & \begin{sideways}pole\end{sideways} & \begin{sideways}traffic sign\end{sideways} \\ 
\midrule
KPConv \cite{thomas2019iccv}/PointPillars \cite{lang2019cvpr} & 52.5  & 88.8 & 72.7 & 61.3 & 31.6 & 90.5 & 72.5 & 17.2 & 9.2 & 30.8 & 19.6 & 84.8 & 69.2 & 69.1 & 29.9 & 59.4 & 22.8 & 64.2 & 56.4 & 47.4\\
RangeNet++ \cite{milioto2019iros}/PointPillars \cite{lang2019cvpr} & 45.9  & 91.8 & 75.1 & 65.0 & 27.7 & 87.4 & 66.9 & 6.7 & 3.1 & 16.2 & 8.8 & 80.5 & 55.1 & 64.8 & 14.6 & 31.8 & 13.5 & 58.6 & 47.9 & 55.9\\
\bottomrule
\end{tabular}}
\caption{Detailed class-wise results of test set results on SemanticKITTI in fixed panoptic quality (PQ$^\dagger$) \cite{porzi2019cvpr}. All results in [\%].}
\label{tab:results_panoptic_fixed_class}
\end{table*}

\emph{Training on KITTI Object Detection: }
Following the original approach of Lang \etal~\cite{lang2019cvpr}, we use $0.16\,$m as voxel grid resolution with a maximum of $12.000$ pillars with at most $100$ points for each pillar for training on the KITTI Object Detection subset of the KITTI Vision Benchmark~\cite{geiger2012cvpr}.
As commonly done and also advocated by Lang \etal~\cite{lang2019cvpr}, we trained a network for \emph{cars}, car network, and a separate network for \emph{pedestrian} and \emph{cyclist}, called pedcyclist network.

For the car network, we consider the part in front of the car inside the ranges $x = (0.0, 69.12)$, $y = (-39.68, 39.68)$, and $z = (-3.0, 1.0)$, where we assume that the sensor is located at $(0,0,0)$.
For the pedcyclist network, we use $x = (0.0, 48.0)$, $y = (-20.0, 20.0)$, and $z = (-2.5, 0.5)$ as volume of the point pillar grid.

\emph{Testing on SemanticKITTI: }
Here, we are interested in predicting bounding boxes for the full field-of-view of the sensor.
Thus, we adapted the parameters for inference.

For the car network, we use a grid volume of size\\ $x=(-69.12,69.12)$, $y=(-69.12, 69.12)$, $z=(-3.0,1.0)$.
For the pedcyclist network, we use a similar grid volume of $x=(-69.12,69.12)$, $y=(-69.12, 69.12)$, $z~=~(-2.5,0.5)$.
Furthermore, we increase the number of maximal pillars to $30000$ and adopt the anchor generation strides to accommodate the large input volume.

\emph{Validation Results on KITTI Detection Benchmark: } \reftab{tab:validation_results} shows the validation set results on the KITTI Detection Benchmark of the trained detector for reference.

\emph{Runtime: } Note that we observed a large discrepancy between the reported runtime and our obtained runtime, which cannot be only explained by using a different system (Nvidia Geforce RTX 2080 Ti vs.~a Nvidia Geforce 1080 Ti).
First, we have to note that the implementations might be different than the originally used implementation.
We believe that the main reason seems to be the $3.4$ times increase input volume and the increased number of pillars.
The fact that the KITTI object detection benchmark only uses a part of the point cloud is also acknowledged in \mbox{Sec\onedot 6} of Lang \etal~\cite{lang2019cvpr}.
We furthermore do not use TensorRT for inference, which could additionally improve the runtime.

\subsection{Baseline Results}

\reftab{tab:panoptic_classes} summarizes the results in a breakdown according to mean Intersection-over-Union (mIoU) and the different panoptic quality metrics.
Due to the overall stronger performance on semantic segmentation of KPConv ($58.8$ mIoU vs. $52.4$ mIoU in \reftab{tab:results_iou_class}),  the panoptic baseline using KPConv is stronger in all metrics.
We believe that this discrepancy can be directly attributed to the stronger performance on small classes.

\reftab{tab:results_panoptic_class}, \reftab{tab:results_iou_class}, and \reftab{tab:results_panoptic_fixed_class} show the detailed results for all classes using the IoU, \ie, just considering the semantic segmentation, panoptic quality and the fixed panoptic quality, respectively.

For the runtime, we assume that the separate object detectors can be run in parallel ($314\,$ ms for \emph{pedestrian/cyclist} and $105\,$ms for \emph{car}) after the semantic segmentation ($200\,$ms  for KPConv and $95\,$ms for RangeNet++) resulting in $514\,$ms and $409\,$ms respectively.

\section{Conclusion}
\label{sec:conclusion}

In this paper, we present an extension of the SemanticKITTI dataset that enables the community to evaluate and benchmark panoptic segmentation approaches using data generated by an automotive LiDAR. We provide the data code as well as an online platform for evaluation using a hidden test set. Additionally, we provide two panoptic segmentation baselines that are built from a combination of state-of-the-art semantic segmentation approaches and a 3D object detector. Ths goal of this dataset paper is to propel the research on LiDAR-based panoptic segmentation and provide a platform for easy benchmarking.

\section*{Acknowledgment}

We thank Hugues Thomas for allowing us to use the predictions from his approach KPConv for our work.

\bibliographystyle{plain}

\bibliography{glorified,new}

\end{document}